\documentclass[final]{cvpr}

\usepackage[accsupp]{axessibility}
\usepackage{times}
\usepackage{epsfig}
\usepackage{graphicx}
\usepackage{amsmath}
\usepackage{amssymb}
\usepackage{mathrsfs}
\usepackage{makecell}
\usepackage{booktabs}
\usepackage{multirow}
\usepackage{caption}
\usepackage{algorithmicx}  
\usepackage{algorithm}
\usepackage{algpseudocode}  
\usepackage{enumitem}
\usepackage{subfig}
\usepackage{bm}
\usepackage{colortbl}
\usepackage{arydshln} 
\usepackage[pagebackref=true,breaklinks=true,colorlinks,bookmarks=false]{hyperref}

\def\cvprPaperID{2004} % *** Enter the CVPR Paper ID here
\def\confYear{CVPR 2022}

\begin{document}

%%%%%%%%% TITLE
%\title{Trash or Treasure? Learning from Noisy Boundary for \\ Semi-Supervised Instance Segmentation}
%\title{Lemon or Lemonade? Learning from Noisy Boundaries for \\ Semi-supervised Instance Segmentation}
\title{Noisy Boundaries: Lemon or Lemonade for Semi-supervised \\ Instance Segmentation?}

\author{
Zhenyu Wang \qquad Yali Li \thanks{Corresponding author} \qquad Shengjin Wang\\

Beijing National Research Center for Information Science and Technology (BNRist)\\
Department of Electronic Engineering, Tsinghua University\\
{\tt\small wangzy20@mails.tsinghua.edu.cn,  \{liyali13, wgsgj\}@tsinghua.edu.cn}

}

\renewcommand{\thefootnote}{}

\maketitle

\pagestyle{empty}
\thispagestyle{empty}

\begin{abstract}

Current instance segmentation methods rely heavily on pixel-level annotated images. The huge cost to obtain such fully-annotated images restricts the dataset scale and limits the performance. In this paper, we formally address semi-supervised instance segmentation, where unlabeled images are employed to boost the performance. We construct a framework for semi-supervised instance segmentation by assigning pixel-level pseudo labels. Under this framework, we point out that \textbf{noisy boundaries} associated with pseudo labels are double-edged. We propose to exploit and resist them in a unified manner simultaneously: 1) To combat the negative effects of noisy boundaries, we propose a noise-tolerant mask head by leveraging low-resolution features. 2) To enhance the positive impacts, we introduce a boundary-preserving map for learning detailed information within boundary-relevant regions. We evaluate our approach by extensive experiments. It behaves extraordinarily, outperforming the supervised baseline by a large margin, more than 6\% on Cityscapes, 7\% on COCO and 4.5\% on BDD100k. On Cityscapes, our method achieves comparable performance by utilizing only 30\% labeled images. \footnote{Codes are available at https://github.com/zhenyuw16/noisyboundaries.}

\end{abstract}

\vspace{-1.0em}

\section{Introduction}

\emph{``When life gives you lemons, make lemonade.''}

\hspace{54mm}\emph{-- \small{Elbert Hubbard}}

%The performance of instance segmentation has been improved significantly with the development of deep learning \cite{he2017mask, huang2019mask, bolya2019yolact, tian2020conditional, wang2020solo}. Although great progress has been made, there remains two problems: 1) Current instance segmentation methods require pixel-level labeled images for fully-supervised training, which are prohibitively expensive to annotate. Statistically, segmenting one object instance requires 79s on average \cite{lin2014microsoft}. In some cases, annotating a single image with high quality even costs more than 1.5h \cite{cordts2016cityscapes}. This severely restricts the scale of current datasets and further limits the performance of existing models. 2) Compared with pixel-level data, unlabeled images are nearly labor-free to acquire and enable us to obtain enormous resources. However, it is hard for current methods to utilize these data. Researches in cognition science \cite{gibson2013human, latourrette2019little} have demonstrated that human concept learning involves large amounts of unlabeled experience without feedback. This motivates us to use unlabeled images to boost the performance of fully-supervised instance segmentation. We call this task \emph{semi-supervised instance segmentation}.

The performance of instance segmentation has been improved significantly with the development of deep learning \cite{he2017mask, huang2019mask, bolya2019yolact, tian2020conditional, wang2020solo}. However, current instance segmentation methods require pixel-level labeled images for fully-supervised training, which are prohibitively expensive to annotate. Statistically, segmenting one object instance requires 79s on average \cite{lin2014microsoft}. In some cases, annotating a single image with high quality even costs more than 1.5h \cite{cordts2016cityscapes}. This severely restricts the scale of datasets and further limits the performance of models. Researches in cognition science \cite{gibson2013human, latourrette2019little} have demonstrated that human concept learning involves large amounts of unlabeled experience without feedback. Works in object detection \cite{sohn2020simple, jeong2019consistency, liu2021unbiased} or semantic segmentation \cite{ouali2020semi, luo2020semi, he2021re} have seeked for semi-supervised learning to alleviate the huge expense of human labeling. However, utilizing nearly labor-free unlabeled images is still unexplored for instance segmentation, partially because of its intrinsic difficulty. These motivate us to use unlabeled images to break through the upper bound of fully-supervised instance segmentation. We call this task \emph{semi-supervised instance segmentation}.

%2) Compared with pixel-level data, unlabeled images are nearly labor-free to acquire and enable us to obtain enormous resources. However, it is hard for current methods to utilize these data.  This motivates us to use unlabeled images to boost the performance of fully-supervised instance segmentation. We call this task \emph{semi-supervised instance segmentation}.

\begin{figure}[t]
   \centering
 \setlength{\abovecaptionskip}{0pt}
 \setlength{\belowcaptionskip}{0pt}
%\subfloat[]{ \includegraphics[width=0.9\columnwidth]{imgs/ssis1.jpg} \label{fig:ssis1}}\\
%\subfloat[]{ \includegraphics[width=\columnwidth]{imgs/ssis2.jpg} \label{fig:ssis2}} \\
%\subfloat[]{ \includegraphics[width=\columnwidth]{imgs/ssis3.jpg} \label{fig:ssis2}} \\
   %\caption{(a): Illustration of semi-supervised instance segmentation, where the model is trained on pixel-level annotated images and unlabeled images. (b): Our proposed semi-supervised method improves the performance of its supervised baseline significantly.}
\includegraphics[width=\columnwidth]{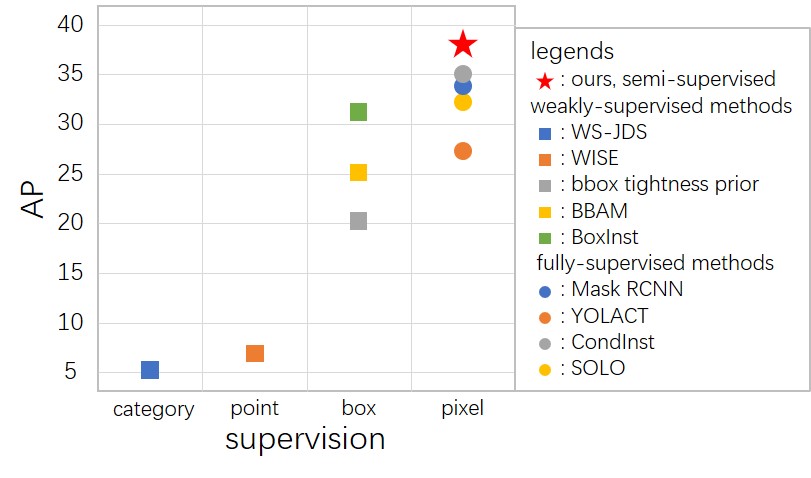}
\caption{Semi-supervised instance segmentation (on the COCO dataset), which explores to utilize unlabeled images, is a novel problem that has not been formally defined and addressed so far. Compared with weakly-supervised and fully-supervised methods, it excavates existing data sufficiently and seeks to use a large number of unlabeled resources, making instance segmentation more practical.}
   \label{fig:ssis}
\end{figure}

%\vspace{-0.4em}

The difficulty to collect pixel-level annotated data in instance segmentation has been recognized by many works. Most of them attempt to deal with this problem by weakly-supervised instance segmentation \cite{hsu2019weakly, tian2021boxinst, shen2019cyclic}. The main benefit of semi-supervised instance segmentation, compared to fully-supervised and weakly-supervised ones, is that it exploits existing resources sufficiently and allows to pursue larger-scale learning. Pixel-level annotated images have been provided in several existing datasets \cite{lin2014microsoft, cordts2016cityscapes, yu2020bdd100k}. Semi-supervised instance segmentation can utilize these data, which are necessary for high-quality segmentation masks. Unlabeled images are enormous, and obtaining them is easy. As a result, the scale of learning is not restricted by datasets and can be as large as possible. This endows semi-supervised instance segmentation the potential to achieve better performance continuously. %, a capability that is critical for practical application. 

Stimulated by the importance of semi-supervised instance segmentation, it is natural to ask: \emph{what's the core issue of semi-supervised instance segmentation?} The core is to excavate information within unlabeled data. Progress in fully-supervised or weakly-supervised methods cannot be applied to the semi-supervised task, as supervision clues are necessary for them. To tackle this issue, we adopt the idea of pseudo labels and propose a semi-supervised instance segmentation framework. With this framework, the noise, especially included in masks from pseudo labels, is essential for the effective exploitation of unlabeled images. Considering that a high proportion of pixel-level noise lies in boundary regions, we focus on noisy boundaries. They provide incorrect supervision signals, but also contain many details that contribute to the model performance. This contradiction makes noisy boundaries a challenging problem. 

%

%With this framework, a problem arises:  Considering that noise in pseudo labels impedes the segmentation performance seriously, we focus on pixel-level noise for the instance segmentation task. In practice, most pixel-level noise lies in boundary regions. We thus conclude that noisy boundaries are crucial to semi-supervised instance segmentation.

%However, noisy boundaries are a complicated problem. 

In a word, noisy boundaries are double-edged (both ``lemon'' and ``lemonade''), including useful and harmful information together.  \emph{How to learn from noisy boundaries for semi-supervised instance segmentation?} We need to exploit and combat them jointly. Specifically, we propose a noise-tolerant mask head (NTM) and a boundary-preserving map (BPM). Our NTM introduces a mask prediction branch for low-resolution segmentation output. With a low-resolution ground-truth for supervision, the details from boundaries are eliminated, where most of the noise exists. This contributes to noise-resistant learning. Meanwhile, our proposed BPM facilitates boundary learning. Different from previous approaches which preserve boundaries at the cost of enlarging pixel-level noise, our BPM strongly corresponds with the boundary regions but is irrelevant to noise. This leads to more precise results. With the help of our NTM and BPM, our method benefits from valuable features within noisy boundaries and discards the detrimental ones, thus mining unlabeled information more effectively.

Our main contributions can be summarized as follows:

\begin{itemize}[topsep=3pt, parsep=0pt, itemsep=3pt, partopsep=0pt]
    \item We formally address the semi-supervised instance segmentation task and construct a framework to exploit unlabeled data, which empowers us ability to break through the fully-supervised upper bound.
    \item We demonstrate the negative correlation between mask resolution and pixel-level noise, then propose a noise-tolerant head by interweaving low and high resolution features, which can resist noise in boundary regions.
    \item We propose a boundary-preserving map, which enriches boundary-relevant regions and suppresses narrow noise-excessive regions simultaneously. This produces more accurate segmentation boundaries.%This contributes to much higher quality segmentation results.
\end{itemize}
 
Extensive experiments on Cityscapes \cite{cordts2016cityscapes}, COCO \cite{lin2014microsoft} and BDD100K \cite{yu2020bdd100k} demonstrate the effectiveness of our method. It obtains comparable results with only 30\% amount of labeled images and surpasses its fully-supervised counterpart with only 40\% labeled data. The performance is even better than approaches using human-annotated coarse labels or extra box-level annotations. We provide a simple and effective framework, which we believe will facilitate future research towards this direction.
 
\section{Related Work}

\textbf{Instance Segmentation.} Most of instance segmentation methods can be categorized into detection-based methods. Mask RCNN \cite{he2017mask} extends Faster RCNN \cite{ren2015faster} to instance segmentation by adding an FCN based mask prediction branch. PANet \cite{liu2018path} introduces bottom-up path augmentation for better feature learning. Cascade Mask RCNN \cite{cai2019cascade} extends Cascade RCNN \cite{Cai_2018_CVPR} to instance segmentation. HTC \cite{chen2019hybrid} further interweaves feature learning and adopts semantic knowledge to facilitate instance segmentation learning. Following works \cite{huang2019mask, kirillov2020pointrend, cheng2020boundary, zhang2021refinemask} continue to improve the performance of instance segmentation. Recently, one-stage methods \cite{bolya2019yolact, tian2020conditional, wang2020solo, chen2020blendmask, wang2020solov2} also develop rapidly and achieve satisfying results with faster speed. They aim to predict masks directly, instead of generating proposals first. However, all of these methods require pixel-level annotated images, which are expensive to obtain.

\textbf{Instance Segmentation with Incomplete Supervision.} Considering the difficulty of obtaining pixel-level annotated images, some recent works aim to use incomplete annotations for instance segmentation. Weakly-supervised methods perform instance segmentation using either box-level labels \cite{hsu2019weakly, tian2021boxinst, lee2021bbam} or image-level tags \cite{shen2019cyclic, fan2018associating}. However, they do not utilize existing pixel-level annotations, thus hard to obtain satisfying results compared to fully-supervised ones. Partially supervised approaches \cite{hu2018learning, kuo2019shapemask, zhou2020learning} adopt the setting where a small number of categories are pixel-level annotated and others have only box-level annotations. They aim to utilize box labels to expand the number of categories. Different from them, our target is to improve the performance of fully-supervised networks by using extra unlabeled data.

\begin{figure*}[]
\centering
   \setlength{\abovecaptionskip}{5pt}
   \setlength{\belowcaptionskip}{0pt}
      \includegraphics[width=\textwidth]{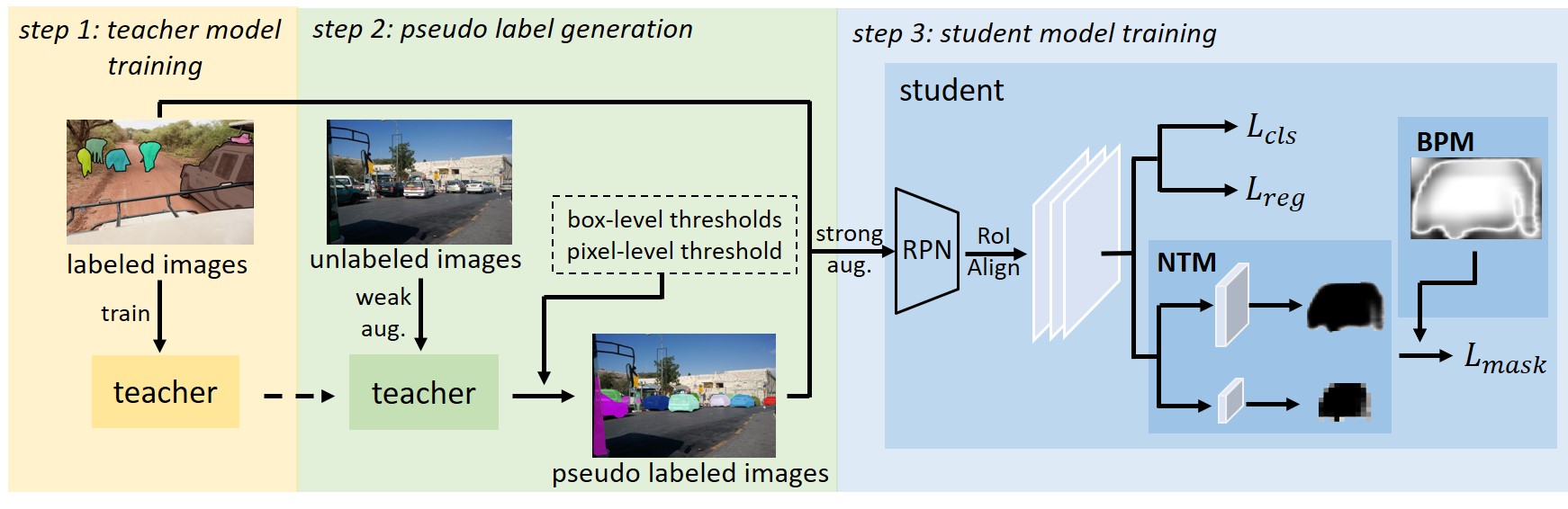}

\caption{\textbf{Framework for semi-supervised instance segmentation.} A teacher model is trained with labeled images, then extracts pseudo labels for unlabeled images. After data aug., these images serve for training the student model. Our noise-tolerant mask head (NTM head) and boundary-preserving map (BPM) helps the student better learn from noisy boundaries.}
\label{fig:bigframe}
\end{figure*}

\textbf{Semi-supervised Learning.} Training models with both labeled and unlabeled images as semi-supervised learning has been widely used in image classification to boost the performance of fully-supervised learning. The prevailing methods include consistency regularization \cite{laine2016temporal, miyato2018virtual, tarvainen2017mean}, pseudo labeling \cite{lee2013pseudo, berthelot2019mixmatch, berthelot2019remixmatch, sohn2020fixmatch}, data augmentation \cite{xie2019unsupervised, berthelot2019mixmatch, sohn2020fixmatch}, or label propagation \cite{zhu2002learning, bengio11label}. Recent works have extended semi-supervised learning to object detection and semantic segmentation. For example, \cite{jeong2019consistency, ouali2020semi, ke2020guided} adopts the idea of consistency regularization and \cite{luo2020semi, liu2021unbiased, tang2021humble, wang2021data, he2021re} utilizes pseudo labels. Recently, self-supervised learning \cite{chen2020simple, chen2020improved, chen2021exploring} also utilizes unlabeled images. The difference is that self-supervised learning trains pretext tasks and is agnostic from downstream tasks, while semi-supervised learning targets at the specific task. In this work, we adopt pseudo labels to solve semi-supervised learning for instance segmentation, a naturally more difficult task.

\section{Method}

\subsection{Semi-supervised Instance Segmentation}

Our goal is to address the semi-supervised instance segmentation task. Specifically, we have a set of pixel-labeled images and aim to utilize easily obtained unlabeled data to boost the performance of instance segmentation.

%Our goal is to address instance segmentation in the semi-supervised setting. Specifically, we are given a set of labeled data $\mathcal{D}_l = \{({\rm x}_i^l, y_i^l)\}_{i=1}^{N_l}$ and a set of unlabeled data $\mathcal{D}_u = \{{\rm x}_i^u\}_{i=1}^{N_u}$, where ${\rm x}$ denotes the given image and $y$ is its corresponding pixel-level ground-truth annotation. We aim to utilize easily obtained unlabeled data $\mathcal{D}_u$ to boost the performance of fully-supervised instance segmentation, which only adopts $\mathcal{D}_l$ in practice.

Our basic framework consists of three steps:

\noindent \textbf{Step 1: Teacher Model Training.} We first train a teacher model with only labeled data  as the common supervised learning. The teacher model will be applied to generate pixel-level pseudo labels for training the student model in the later steps. We choose Mask RCNN \cite{he2017mask} as our teacher model but do not restrict to it.

%This process is exactly the same as the usual supervised learning.

\noindent \textbf{Step 2: Pseudo Label Generation.} With the pre-trained teacher model, we perform inference on unlabeled images to produce instance segmentation masks. Data augmentation as scaling and horizontal flipping is conducted to improve the mask quality and reduce the miscalibration of neural networks \cite{ayhan2018test}. We refer to this as weak augmentation.

To acquire pseudo labels, the raw inference masks need to be processed by two kinds of thresholds: the box-level and the pixel-level ones. At the box level, a large number of bounding boxes are predicted to guarantee a high recall. We thus need to filter low-quality boxes with a confidence threshold. At the pixel level, instance segmentation methods usually calculate foreground probability with \emph{sigmoid}. A probability threshold is required to separate foreground and background pixels for creating masks.

Existing methods usually set the thresholds in a straightforward way. The box-level threshold is usually fixed to 0.7 or 0.9 \cite{sohn2020simple, liu2021unbiased, tang2021humble}, and the pixel-level threshold is generally taken as 0.5. However, this setting way is inappropriate. For the detection branch, current models with \emph{softmax} for category probability are prone to be biased and predict the dominant classes. For the mask branch, the imbalance between foreground and background pixels also affects the prediction. In such a situation, a single threshold is easy to amplify the imbalance problem in pseudo labels.

%Therefore, we follow \cite{radosavovic2018data} and adopt a per-category threshold. For each category, the threshold is set to keep the average number of instances per image the same for the labeled dataset and unlabeled dataset. 

To tackle this issue, we set the thresholds to match the distribution between labeled and unlabeled images. At the box level, we follow \cite{radosavovic2018data} and apply a per-category threshold. For each category, the principle is to keep the average number of instances per image matched for the labeled and unlabeled dataset. Similarly, after filtering low-quality boxes, we set the pixel-level threshold to keep the ratio of foreground to background pixels equivalent. Since mask prediction is acted for RoIs, we only count pixels in the bounding boxes. Also, this threshold is class-agnostic since mask and class prediction is usually decoupled. Note that in the test phase, we still adopt the 0.5 threshold, as we cannot access the distribution of test datasets.

\noindent \textbf{Step 3: Student Model Training.} With thresholding at the box-level and pixel-level, we obtain pseudo labels with mask annotations (pseudo masks). They are treated as the ground-truth labels for training the student model. According to previous works on semi-supervised learning, the diversity of the student model is crucial \cite{tarvainen2017mean} and data augmentation is important \cite{sohn2020fixmatch, sohn2020simple, liu2021unbiased}. We thus conduct data augmentation for images when training the student model, mainly including color transformation and cutout. We call this strong augmentation, to differ from that in the pseudo label generation step. Note that we do not adopt any augmentation strategy in the test phase for a fair comparison.

\subsection{Noise-tolerant Mask Head}

The above framework enables us to train a semi-supervised instance segmentation model. However, noise inherently exists in pseudo masks, which impedes the performance. We need a noise-resistant learning to combat it.

When training the student model, each proposal generated by RPN \cite{ren2015faster} will be assigned a mask from pseudo labels. After RoI-Align and a mask head, the mask prediction is generated. The assigned mask supervises this learning process. In the pseudo mask circumstance, assigned labels are not always accurate. The incorrect labels mislead the learning and deteriorate the performance. We design a noise-tolerant mask head (NTM) to help our model better resist noise in pseudo labels.

\begin{figure}[]
\subfloat[]{ \includegraphics[width=0.48\columnwidth]{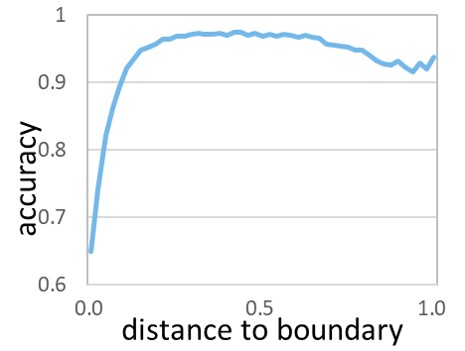} \label{fig:21}  }
     %\hspace{0.04\columnwidth}
\raggedright \subfloat[]{ \includegraphics[width=0.48\columnwidth]{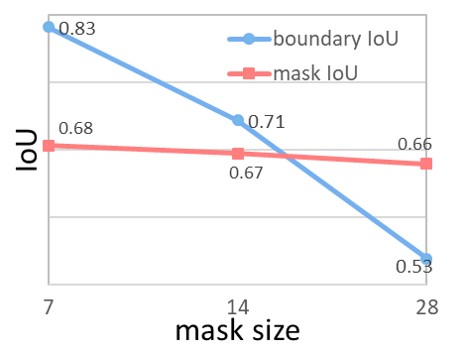} \label{fig:22} }
\caption{ \textbf{Illustration of pixel-level noise in instance segmentation.} (a): the mean accuracy of pixels \emph{v.s.} their relative distance to the boundary. (b): the mean IoU between pseudo mask labels and their ground-truth ones \emph{v.s.} their sizes. Pixels that are closer to the boundary are more likely to be noisy, and reducing size can suppress noise.}
\label{fig:ofd}
\end{figure}

To resist noise in the learning process, we need to investigate which pixels are more likely to be noisy. The answer is: pixels that are closer to the boundaries. Boundary-relevant regions are noisier because they usually correspond with the decision boundary, where category features are not salient. Also, they contain detailed information that is difficult to learn. To further verify this, we conduct an empirical study on the Cityscapes dataset \cite{cordts2016cityscapes} and plot it in Fig. \ref{fig:21}. The mean accuracy of pixels is high, more than 90\%. However, for pixels that are extremely close to the boundaries, the mean accuracy is significantly lower. Therefore, to resist noise in pseudo masks, the key lies in boundary-related regions. Their details and features are only visible when the mask resolution is high enough. In Mask RCNN \cite{he2017mask}, the mask ground-truth is usually downsized to $28 \times 28$. If the size turns smaller and the resolution is lowered, the image details can be implicit, where noise mainly lies. This is demonstrated in numerical analysis from Fig. \ref{fig:22}. As the mask size decreases, the overall mask IoU between pseudo labels and their corresponding ground-truth labels increases a bit, and the boundary IoU \cite{cheng2021boundary} improves significantly. We conclude that downsizing masks benefits the quality of pseudo labels, especially for regions near boundaries. 

%This resolution already reserves details for most instances. As a result, noise is reduced. 

\begin{figure}[t]
   \centering
   \setlength{\abovecaptionskip}{5pt}
   \setlength{\belowcaptionskip}{5pt}
   \includegraphics[width=\columnwidth]{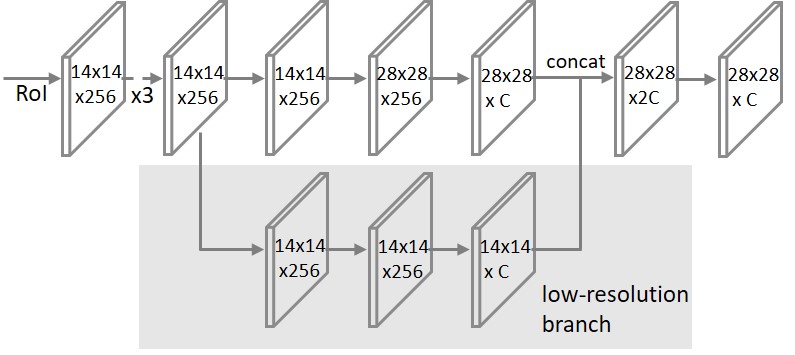}
   \caption{\textbf{The structure of our noise-tolerant mask head.} We add a branch for predicting results with low resolution. The low resolution mask better resists noise thus makes the network more noise-tolerant. Arrows, unless otherwise specified, denote conv or deconv layers. The conv kernel sizes are all the same as that in Mask RCNN. C denotes the number of categories.}
   \label{fig:ntmh}
\end{figure}

Motivated by the above analysis, we propose our noise-tolerant mask head. We add a branch for the low-resolution mask prediction, and the structure is illustrated in Fig \ref{fig:ntmh}. This branch is supervised by a smaller size mask (we adopt 14 in practice). With a small size and low resolution, its features are cleaner and more noise-resistant. Consequently, this branch is able to utilize more accurate information, which contributes to learning in the semi-supervised setting. However, since the resolution is low, the predicted segmentation results are coarse and hard to reserve details. Therefore, the original high-resolution mask head is still retained. Specifically, the original high-resolution branch aims to learn fine-grained information, which is more likely to be affected by noise, while the low-resolution branch targets at learning coarse but clean information. Features from the low-resolution branch are fused into the high-resolution branch to pass clean messages. With this structure, we achieve more noise-tolerant learning. In the test phase, we only apply the high-resolution branch.

\subsection{Boundary-preserving Map}

With the noise-tolerant mask head, our model better resists noise from boundary-related regions. However, boundaries are also essential for instance segmentation, since detailed information within them is necessary for the quality of the predicted masks. In this subsection, we propose a boundary-preserving map (BPM) to assist boundary learning in the semi-supervised task.

\begin{figure}[t]
   \centering
   \setlength{\abovecaptionskip}{5pt}
   \setlength{\belowcaptionskip}{5pt}
   \includegraphics[width=\columnwidth]{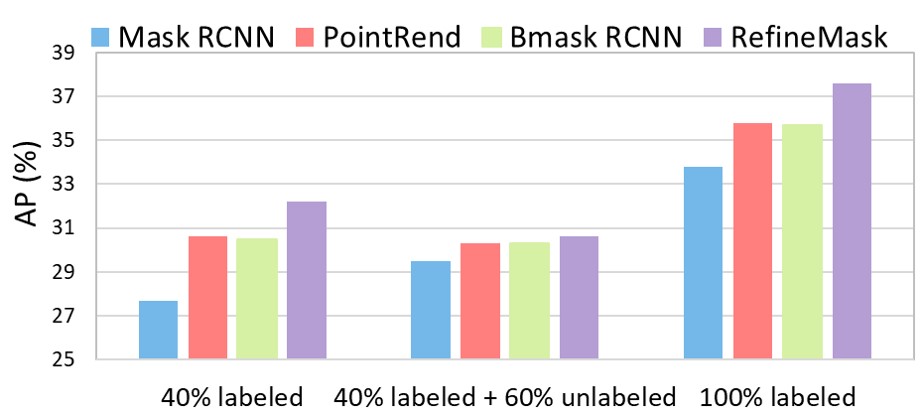}
   \caption{\textbf{Performance of existing boundary-preserving models on fully-supervised and semi-supervised tasks.} These methods improve the performance significantly for the fully-supervised task, but is limited in the semi-supervised setting because of the noise in boundary regions.}
   \label{fig:bmodel}
\end{figure}

Facilitating boundary learning has been discussed in recent works such as PointRend \cite{kirillov2020pointrend}, BMask RCNN \cite{cheng2020boundary} and RefineMask \cite{zhang2021refinemask}. These methods are effective for fully-supervised learning but limited to the semi-supervised task. To corroborate this, we perform experiments on the Cityscapes dataset with 40\% randomly selected images as labeled ones, and plot the acquired mask $AP$ in Fig. \ref{fig:bmodel}. It is observed that these methods improve more than 2\% compared to the Mask RCNN baseline in the fully-supervised task, but less than 1\% for semi-supervised learning. The reason lies in noisy boundaries. In the semi-supervised task, traditional methods promote boundary learning at the cost of increasing the adverse effects of boundary-aware noise. Consequently, these methods are not suitable for semi-supervised segmentation.
%For example, PointRend \cite{kirillov2020pointrend} produces predictions of varying resolution thus improves the boundary segmentation ability, BMask RCNN \cite{cheng2020boundary} adds a boundary supervision loss to promote boundary learning, and RefineMask \cite{zhang2021refinemask} utilizes semantic knowledge and boundary-aware refinement for higher resolution. 

In semi-supervised learning, the importance is thus to preserve boundaries but not to amplify  noise. To promote boundary learning, the model should focus more on pixels that are closer to the boundaries. To reduce noise, pixels that are most likely to be noisy should be suppressed during training. Fig. \ref{fig:21} indicates that noise is excessive for those pixels whose distances to boundaries are extremely small. So we need to repress these pixels. Based on the above analysis, we present our boundary-preserving map. In BPM, the value of a pixel is negatively correlated with its distance to the boundaries. The only exception is pixels that are extremely close to boundaries, whose values should be small to suppress noise. Distance calculation for all pixels is effective but computationally complex, significantly decreasing the training speed. Denote the mask probability output by \emph{sigmoid} function as ${\bm p}=[p_{ij}]$. We find that the laplace operation of the probability map, $\nabla^2{\bm p}$, well meets the above requirement and is computationally efficient. As a result,  we adopt $\nabla^2{\bm p}$ as our BPM. We directly use our BPM to re-weight mask loss for different pixels, which is a simple but effective strategy. 

\begin{figure}[t]
   \centering
   \setlength{\abovecaptionskip}{5pt}
   \setlength{\belowcaptionskip}{5pt}
      \subfloat[]{ \includegraphics[width=0.8\columnwidth]{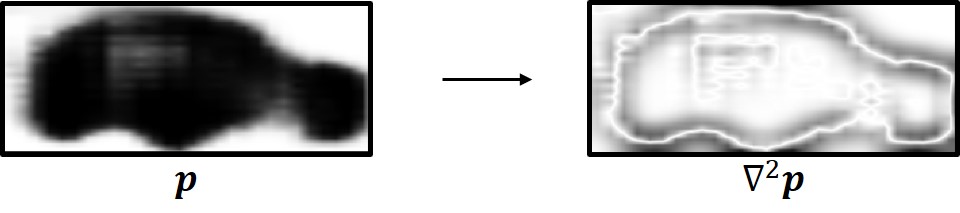} \label{fig:deltap1}}\\
      \subfloat[]{ \includegraphics[height=0.28\columnwidth]{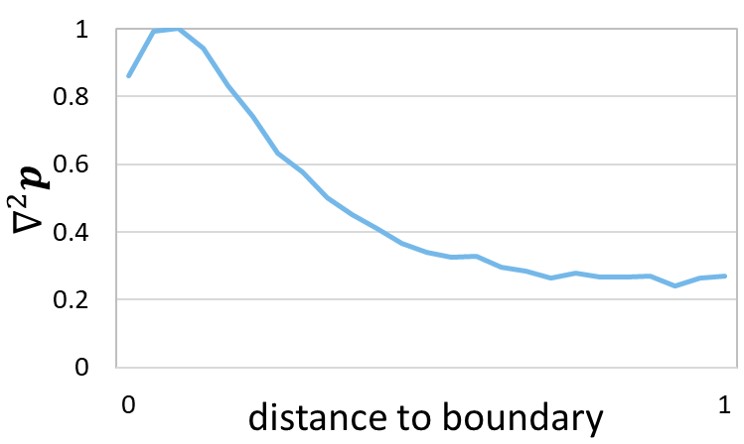} \label{fig:deltap2}}
      \subfloat[]{ \includegraphics[height=0.28\columnwidth]{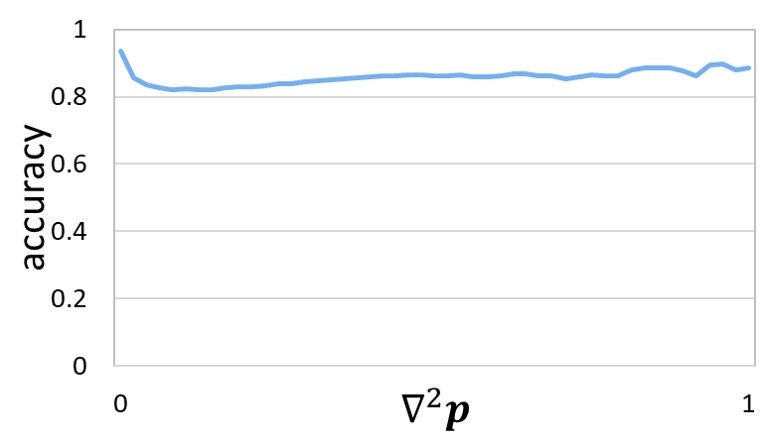} \label{fig:deltap3}}
      \\
   \caption{\textbf{Illustration for our boundary-preserving re-weighting.} (a): illustrative examples, (b): $\nabla^2{\bm p}$ \emph{v.s.} the distance of the pixel to the boundary, (c): the mean of accuracy of pixels \emph{v.s.} their values of $\nabla^2{\bm p}$.}
   \label{fig:deltap}
\end{figure}

We show illustrative examples of our BPM in Fig. \ref{fig:deltap}. For pixels that belong to boundary-relevant regions but do not lie in the narrow band along the boundaries, their values are the highest. These pixels usually contain detailed information and are relatively clean, thus should be paid attention to. Also, because of this design, our BPM is somewhat irrelevant to noise. With this property, our BPM benefits boundary learning and does not increase the effect of noise. This makes it appropriate for the semi-supervised task.

\section{Experiments}

We evaluate our proposed method on Cityscapes \cite{cordts2016cityscapes}, MS COCO \cite{lin2014microsoft} and BDD100K \cite{yu2020bdd100k}. Cityscapes provides 2,975 images for the training set. Besides, it consists of 20,000 images with coarse annotations. COCO includes 118,287 images. It also provides 123,403 unlabeled images. BDD100K is a diverse dataset about visual driving scenes. Only a subset of BDD100K is pixel-level annotated: about 7k images with mask annotations and 70k images with box annotations. Among them, 67k images have box-level annotations but no pixel-level labels. Our method is implemented with Pytorch \cite{paszke2019pytorch} and MMDetection \cite{chen2019mmdetection}. Unless otherwise specified, we use Mask RCNN \cite{he2017mask} with ResNet50 \cite{he2016deep} and FPN \cite{lin2017feature}.

\subsection{Experiments on Cityscapes}

\begin{table}[]
\centering
\setlength{\abovecaptionskip}{0pt}
\setlength{\belowcaptionskip}{0pt}
\caption{\textbf{Experimental Results on Cityscapes with a varying percentage of labeled images.} $\dagger$ denotes adopting the same data augmentation in the semi-supervised training. $^\S$ denotes using focal loss for the detection branch.}
\begin{tabular}{c|ccccc}
\Xhline{1.2pt} 
Method & 5\%  & 10\%  & 20\%  & 30\%  & 40\% \\
\hline

supervised &   11.8 & 16.8 & 22.3 & 26.3 & 27.7\\
supervised $\dagger$ & 11.3 & 16.4 & 22.6 & 26.6 & 28.3\\
\hline
\multicolumn{6}{c}{\emph{semi-supervised object detection methods}} \\
\hline
DD \cite{radosavovic2018data} & 13.7 & 19.2 & 24.6 & 27.4 & 29.5\\
STAC \cite{sohn2020simple} & 11.9 & 18.2 & 22.9 & 29.0 & 29.8\\
CSD \cite{jeong2019consistency} & 14.1 & 17.9 & 24.6 & 27.5 & 28.9\\
Ubteacher \cite{liu2021unbiased} & 16.0 & 20.0 & 27.1 & 28.0 & 29.6\\
\hline
\multicolumn{6}{c}{\emph{semi-supervised semantic segmentation methods}} \\
\hline
CCT \cite{ouali2020semi} & 15.2 & 18.6 & 24.7 & 26.5 & 28.4\\
Dual-branch \cite{luo2020semi} & 13.9 & 18.9 & 24.0 & 28.9 & 28.9\\
\hline
\multicolumn{6}{c}{\emph{semi-supervised instance segmentation methods}} \\
\hline
baseline & 15.7 & 20.2 & 25.5 & 28.3 & 29.5\\
ours & \textbf{17.1} & \textbf{22.1} & \textbf{29.0} & \textbf{32.4} & \textbf{33.0}\\
ours $^\S$ & \textbf{21.2} & \textbf{23.7} & \textbf{30.8} & \textbf{33.2} & \textbf{34.1}\\
\Xhline{1.2pt} 
\end{tabular}
%}
\label{tab:cityscapespercent}
\end{table}

%\textbf{Experiment and baseline settings.} We evaluation our method on the cityscapes validation set. We randomly select a certain percentage of images from the training set as labeled images and treat the rest as unlabeled ones. The pseudo label method without data augmentation, NTM and BPM, is our semi-supervised baseline. Besides, we conduct experiment with all training set images, and utilize the extra coarse-annotated images as unlabeled ones. We design the following experiments for this setting comparison. Supervised: only labeled images; coarse GT: directly using the given coarse annotations; coarse finetune: firstly training with coarse-annotated images, then finetuning with the pixel-level images; fine $\rightarrow$ coarse $\rightarrow$ fine: firstly training the model with pixel-level annotated images, then finetuning with coarse-annotated images, finally finetuning with pixel-level annotated images, just as \cite{yuan2020object}.

\textbf{Experiments with a varying percentage of labeled images.} We evaluate our method on the Cityscapes validation set. We randomly select a certain percentage of images from the training set as labeled images and treat the rest as unlabeled ones. Since semi-supervised instance segmentation is a new task, we extend methods about the two most relevant tasks - semi-supervised object detection and semantic segmentation for comparison. Results from  Tab. \ref{tab:cityscapespercent} suggest that these methods are strong baselines. The pseudo label method without data augmentation, NTM and BPM, is our semi-supervised instance segmentation baseline.

The results are listed in Tab. \ref{tab:cityscapespercent}. Our approach behaves consistently better under different degrees of supervised data. Compared to unbiased teacher \cite{liu2021unbiased}, the state-of-the-art detector in semi-supervised object detection, our method outperforms it by a large margin under various settings. When the labeled ratio is 30\% and 40\%, the $AP$ improvement reaches 4.4\% and 3.4\%. For CCT \cite{ouali2020semi}, one recent effective semi-supervised semantic segmentation method, our approach surpasses it by almost 6\% in the 30\% setting.  Compared with the semi-supervised instance segmentation baseline, our method basically enhances the mask $AP$ by 3\%. When the labeled ratio is moderate, the increase is more: 3.5\% and 4.1\% for the 20\% and 30\% labeled ratio respectively. This demonstrates that our method learns better from noisy boundaries. Compared with the supervised counterpart, we achieve a more than 6\% improvement. The importance of unlabeled images and the necessity of semi-supervised instance segmentation is validated.

Our method aims to learn from noisy boundaries for semi-supervised instance segmentation, thus targeting at the mask prediction branch. Focal loss \cite{lin2017focal} has been proved beneficial to semi-supervised object detection. We apply focal loss for the detection branch and the segmentation accuracy can be further improved. In particular, when the labeled percentage is 40\%, the mask $AP$ is 34.1\%, which is higher than the fully-supervised method where all images are pixel-annotated (33.8\%). When labeled images are 30\%, the 33.2\% $AP$ is also comparable. This substantiates the great potential of semi-supervised learning.
%, we observe that the semi-supervised instance segmentation baseline basically improves the performance of the supervised method by more than 2\%, especially when labeled ratio is small. The importance of unlabeled images and the necessity of semi-supervised instance segmentation is thus validated  The superiority remains obvious when the same data augmentation is adopted. 

\begin{table}[]
\centering
\setlength{\abovecaptionskip}{0pt}
\setlength{\belowcaptionskip}{0pt}
\caption{\textbf{Experimental Results on Cityscapes with coarse-annotated images.} $\dagger$ denotes adopting the same data augmentation in the semi-supervised training. $^\S$ denotes using focal loss for the detection branch.}
\begin{tabular}{c|ccc}
\Xhline{1.2pt} 
Method & $AP$ & $AP_{50}$ & $AP_{75}$\\
\hline
supervised &  33.8 & 61.8 & 31.4\\
supervised $\dagger$ & 34.7 & 61.8 & 33.7\\
coarse GT & 23.3 & 49.4 & 18.3\\
coarse finetune & 34.2 & 59.9 & 32.3\\
fine $\rightarrow$ coarse $\rightarrow$ fine & 35.8 & 62.9 & 35.3\\
ours & \textbf{39.3} & \textbf{65.6} & \textbf{38.9}\\
ours $^\S$ & \textbf{41.1} & \textbf{68.2} & \textbf{42.1}\\
\Xhline{1.2pt} 
\end{tabular}
%}
\label{tab:cityscapescoarse}
\end{table}

\textbf{Experiments with coarse-annotated images.} We also conduct experiments with all images from the training set and utilize the extra coarse-annotated images as unlabeled ones. We design the following experiments for comparison. Coarse GT: directly using the given coarse annotations; coarse finetune: firstly training with coarse-annotated images, then finetuning with fine-annotated images; fine $\rightarrow$ coarse $\rightarrow$ fine: firstly training the model with  fine-annotated images, then learning with coarse-annotated images, finally finetuning with  fine-annotated images, just as \cite{yuan2020object}.

From Tab. \ref{tab:cityscapescoarse}, we notice that with the 20,000 images, our method achieves the 39.3\% $AP$, which exceeds supervised learning by 5.5\%. This demonstrates that our semi-supervised method helps the model get rid of the limitation of the dataset scale. Our method behaves better than designed approaches using human-labeled coarse annotations - more than 3.5\% higher than them. \textbf{Utilizing unlabeled images obtain even better performance than using human-labeled images!} This proves the high effectiveness of our method to exploit unlabeled information. With focal loss, we obtain a 41.1\% $AP$. This remarkable performance corroborates the capability of semi-supervised instance segmentation for practical application.

\subsection{Ablation Study}

We perform ablation study on Cityscapes using 30\% percent of images as labeled ones. The results are in Tab. \ref{tab:ablation}. We adopt the general mask $AP$ and the boundary $AP$ \cite{cheng2021boundary} to evaluate the quality of masks and boundaries separately. 

\begin{figure*}[]
\centering
   \setlength{\abovecaptionskip}{5pt}
   \setlength{\belowcaptionskip}{0pt}
      \includegraphics[width=0.93\textwidth]{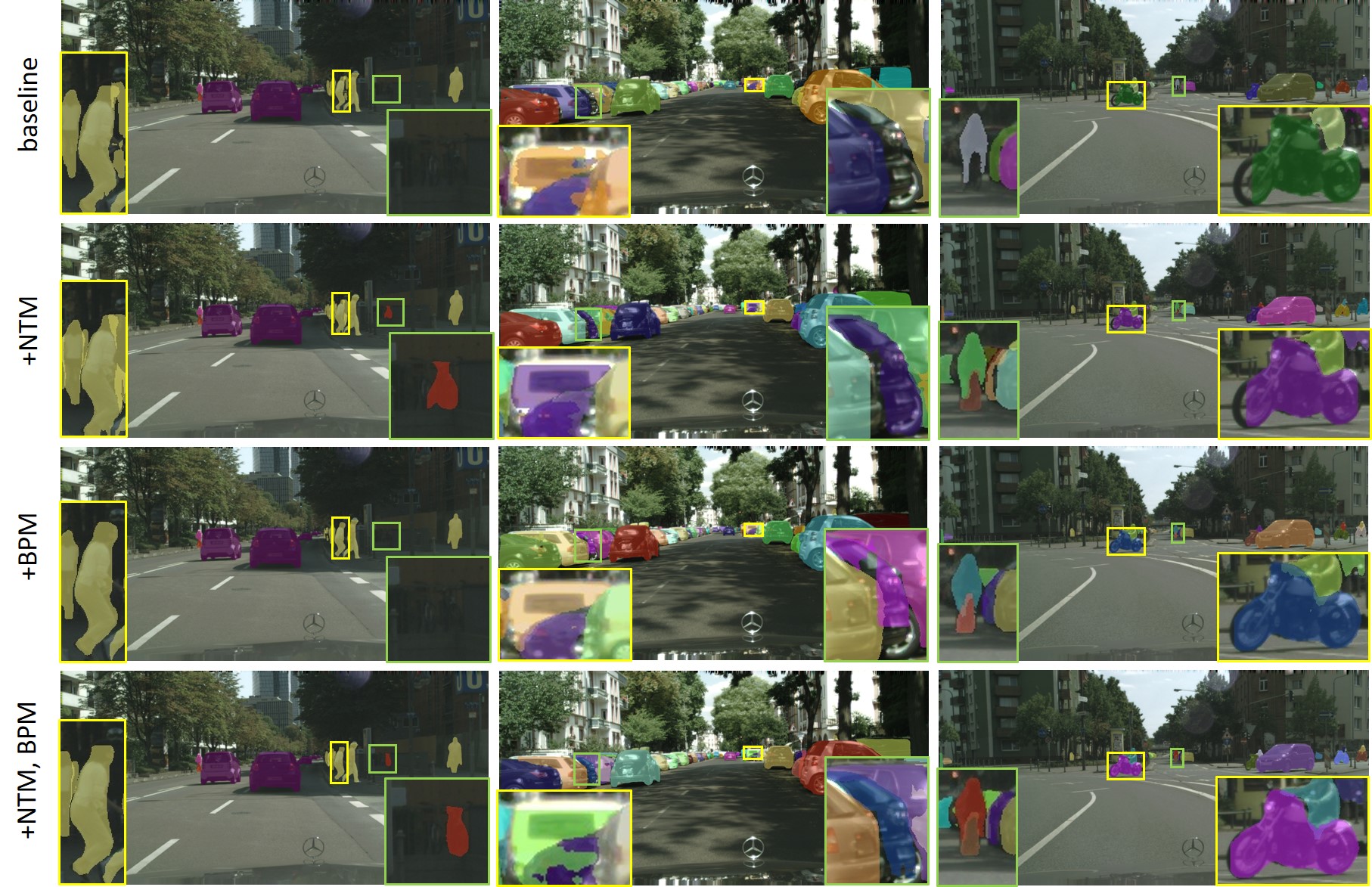}

\caption{\textbf{Illustrative results on Cityscapes to show the effectiveness of our NTM and BPM.} NTM helps more correct detected instances (zoomed in green boxes) and BPM helps more precise boundary (zoomed in yellow boxes).}
\label{fig:comparison}
\end{figure*}

\begin{table}[]
\centering
\setlength{\abovecaptionskip}{0pt}
\setlength{\belowcaptionskip}{0pt}
\caption{\textbf{Ablation study on Cityscapes.} DA: data augmentation,, NTM: noise-tolerant mask head, BPM: boundary-preserving map. We evaluate the mask $AP$ and boundary $AP$, abbreviated as $AP_{bd}$. }
\resizebox{\columnwidth}{!}{
\begin{tabular}{c|ccc|cc}
\Xhline{1.2pt} 
annotations & DA & NTM & BPM & $AP$ & $AP_{bd}$\\
\hline
\multirow{2}{*}{30\% labeled} & & &  & 26.3 & 8.2\\
 & \checkmark & & & 26.6 & 7.8\\
 \hline
\multirow{5}{*}{\shortstack{30\% labeled \\ 70\% unlabeled}} &  &  &  & 28.3 & 10.0\\ & \checkmark & & & 30.2 & 10.4 \\
 & \checkmark & \checkmark & & 31.1 & 10.9\\
 &  \checkmark &  & \checkmark & 31.0 & 11.6\\
 &\checkmark & \checkmark & \checkmark & 32.4 & 11.6\\
\hline
\multirow{2}{*}{100\% labeled} & & &  & 33.8 & 12.7\\
 & \checkmark & & & 34.7 & 12.9\\
\Xhline{1.2pt} 
\end{tabular}
}
\label{tab:ablation}
\end{table}

\textbf{Data augmentation.} We first evaluate the effect of data augmentation in the student model training step. Data augmentation increases the diversity of input samples, hence helping improve the performance. However, it is limited in fully-supervised learning, only bringing a 0.3\% improvement. Even when images are all labeled, the $AP$ increase is still less than 1\%. In comparison, for semi-supervised learning, data augmentation increases the segmentation $AP$ from 28.3\% to 30.2\%. The improvement is more significant, almost 2\%. This corresponds with the conclusion in previous works \cite{sohn2020fixmatch, sohn2020simple, liu2021unbiased} that data augmentation is crucial for the student model in semi-supervised learning.

\textbf{Noise-tolerant mask head.} Results in Tab. \ref{tab:ablation} show that our NTM helps improve the mask $AP$ by 0.9\%, while the boundary $AP$ improvement is not salient. This indicates that NTM helps semi-supervised learning mainly because it benefits the overall segmentation performance, like more correct detected instances or the holistic masks. Since noise in pseudo labels misleads the network learning, the overall discrimination ability of the network is hurt. Our NTM alleviates this problem, thus contributing to the mask $AP$. Illustrative results in Fig. \ref{fig:comparison} also confirm our analysis. In the first and the third images, the missed bicycles are segmented because of the NTM. In the second image, the middle car is detected. The visualized results correspond with the numerical results and our analysis above.

\textbf{Boundary-preserving map.} From Tab. \ref{tab:ablation}, we observe that the mask $AP$ is boosted by 0.8\% with the help of our BPM. Different from NTM, BPM also improves the boundary $AP$ significantly, from 10.9\% to 11.6\%. This indicates that BPM helps the performance mainly because it assists in the quality of boundaries. Such fact is strongly related to the function of BPM: it helps the model focus on boundary regions and learn more detailed information. This is also confirmed by illustrative results in Fig. \ref{fig:comparison}. In the first image, with BPM, the contour of the person, especially at the head part, is more realistic. The same thing also occurs at the top part of the car in the second image and the front wheel of the motor in the third image. This demonstrates the effectiveness of our BPM to boundary learning.

\begin{figure*}[]
\centering
   \setlength{\abovecaptionskip}{5pt}
   \setlength{\belowcaptionskip}{0pt}
      \includegraphics[width=0.9\textwidth]{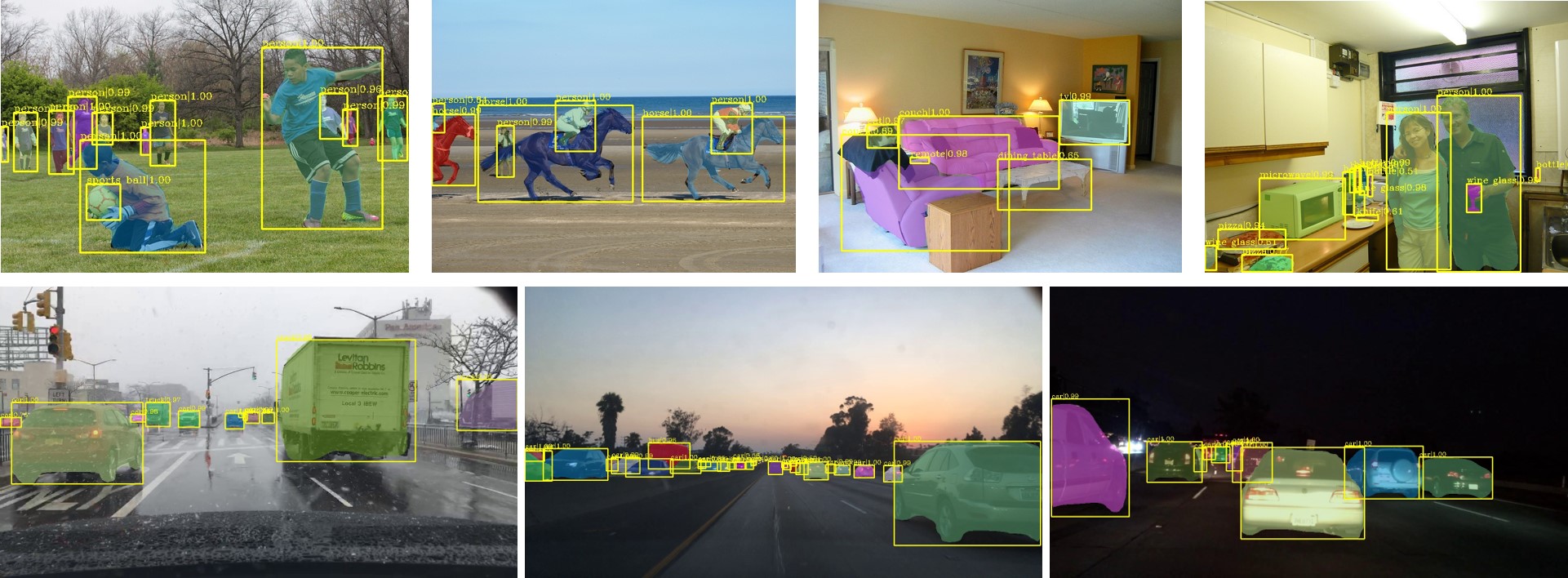}

\caption{\textbf{Instance segmentation results of our method} on COCO (the first row) and BDD100K (the second row).}
\label{fig:ilresults}
\end{figure*}

\subsection{Experiments on COCO}

\begin{table}[]
\centering
\setlength{\abovecaptionskip}{0pt}
\setlength{\belowcaptionskip}{0pt}
\caption{\textbf{Results on COCO with a varying percentage of labeled images.} $\dagger$ denotes data augmentation. We use COCO 120k unlabeled images for the 100\% experiment.}
\resizebox{\columnwidth}{!}{
\begin{tabular}{c|cccccc}
\Xhline{1.2pt} 
Method & 1\%  & 2\%  & 5\%  & 10\% & 30\% & 100\%\\
\hline
supervised & 3.5 & 9.4 & 17.3 & 22.0 & 28.9 & 34.5 \\
supervised $\dagger$ & 3.5 & 9.5 & 17.4 & 21.9 & 29.0 & 37.1\\
DD \cite{radosavovic2018data} & 3.8 & 11.8 & 20.4 & 24.2 & 30.5 & 35.7 \\
ours & \textbf{7.7} & \textbf{16.3} & \textbf{24.9} & \textbf{29.2} & \textbf{32.8} & \textbf{38.6}\\
\Xhline{1.2pt} 
\end{tabular}
}
\label{tab:cocopercent}
\end{table}

%\textbf{Experiments with a varying percentage of labeled images.} 
We also perform experiments on the challenging COCO dataset. Similarly, we randomly select a certain ratio of images from the COCO 2017 training set as labeled data, and the rest of them are unlabeled ones. For the 30\% setting, we simply use the 35k subset of the COCO 2014 validation set as labeled images. For the 100\% setting, we use all images from the COCO 2017 training set as labeled ones and 120k COCO unlabeled images as unlabeled ones. We list the mask $AP$ in Tab. \ref{tab:cocopercent}. Our method continues to be better than the supervised baseline. When the labeled images are 5\% and 10\%, our semi-supervised learning boosts the supervised learning by more than 7\%, which is quite prominent. For the 100\% experiment, where all pixel-annotated images are adopted and utilized data are more, our method achieves a 38.6\% $AP$. The above experiments verify the value of our semi-supervised instance segmentation.

\subsection{Experiments on BDD100K}

\begin{table}[]
\centering
\setlength{\abovecaptionskip}{0pt}
\setlength{\belowcaptionskip}{0pt}
\caption{\textbf{Experimental Results on BDD100K.} }
\resizebox{\columnwidth}{!}{
\begin{tabular}{c|c|ccc}
\Xhline{1.2pt} 
annotations & Method &$AP$ \\
\hline
7k w/ masks & Mask RCNN & 21.6 \\
\hline
\multirow{4}{*}{\shortstack{7k w/ masks\\ 67k w/ boxes}} & Mask RCNN & 24.5 \\  % \multirow{4}{2.4cm}{ 7k w/ masks  67k w/ boxes}
 & Grabcut Mask RCNN  \cite{khoreva2017simple} & 21.0 \\
 & Progressive Mask RCNN \cite{zhou2020learning} & 24.8\\
 & Mask RCNN w/ ShapeProp \cite{zhou2020learning} & 26.2\\
\hline
\multirow{2}{*}{\shortstack{7k w/ masks\\ 67k w/o labels}} & semi baseline & 24.4 \\
 & ours & \textbf{26.3}\\
\Xhline{1.2pt} 
\end{tabular}
}
\label{tab:bdd100k}
\end{table}

We further benchmark on the BDD100K dataset. We use the 7k images with mask annotations as labeled images and the 67k images with only box-level annotations as unlabeled ones. Their box annotations do not participate in training. We compare our results with methods in \cite{zhou2020learning}, where box annotations for the 67k images are utilized for the partially-supervised learning. As Tab. \ref{tab:bdd100k} shows, our method obtains a 26.3\% $AP$, which outperforms its supervised baseline by 4.7\%. Our method performs better than Mask RCNN w/ ShapeProp \cite{zhou2020learning}, where box annotations are utilized. This further indicates that our method utilizes the information within unlabeled images quite sufficiently, so that utilizing unlabeled images outperform previous approaches that adopt box-level annotated images.

\section{Conclusion}

Considering the huge expense of labeling mask annotations, we propose the semi-supervised instance segmentation task. It enables the model to fully excavate available information and explore more extensive resources. With pseudo labels, unlabeled images participate in training and help improve the performance. By further learning from noisy boundaries, we alleviate the negative effects brought by noisy pseudo labels and exploit more valuable information within boundary-relevant regions. The extraordinary performance on benchmark datasets demonstrates the great ability of our method. Semi-supervised instance segmentation is a challenging but interesting problem. We hope that our simple and effective framework will stimulate future research along this direction. %We will explore more efficient way to utilize unlabeled images and larger-scale unlabeled images to facilitate model learning.  

\section*{Acknowledgement}

This work was supported by the state key development program in 14th Five-Year under Grant Nos.2021QY1702, 2021YFF0602103, 2021YFF0602102. We also thank for the research fund under Grant No. 2019GQG0001 from the Institute for Guo Qiang, Tsinghua University.

{\small
\bibliographystyle{ieee_fullname}
\bibliography{egbib}
}

\end{document}